\documentclass[sigconf]{acmart}
\usepackage{dsfont,amsthm,url,mathtools,algorithmic,algorithm, bbold}
\usepackage{subcaption, mwe,epsfig,graphicx,color,booktabs}
\theoremstyle{definition}

\newcommand{\hanwei}[1]{{\color{black}{{#1}}}}
\newcommand{\minus}{\scalebox{0.75}[1.0]{$-$}}
\DeclareMathOperator*{\argmax}{argmax}

\copyrightyear{2020}
\acmYear{2020}
\setcopyright{acmlicensed}\acmConference[ICAIF '20]{ACM International Conference on AI in Finance}{October 15--16, 2020}{New York, NY, USA}
\acmBooktitle{ACM International Conference on AI in Finance (ICAIF '20), October 15--16, 2020, New York, NY, USA}
\acmPrice{15.00}
\acmDOI{10.1145/3383455.3422550}
\acmISBN{978-1-4503-7584-9/20/10}

\begin{document}
\title{Conditional Mutual Information-Based Contrastive Loss \\for Financial Time Series Forecasting}
\author{Hanwei Wu}
\affiliation{%
 \institution{KTH Royal Institute of Technology}
 \city{Stockholm}
  \country{Sweden}
}
\email{hanwei@kth.se}
\author{Ather Gattami}
\affiliation{%
  \institution{RISE Research Institutes of Sweden}
  \city{Stockholm}
  \country{Sweden}
}
\email{ather.gattami@ri.se}
\author{Markus Flierl}
\affiliation{%
  \institution{KTH Royal Institute of Technology}
  \city{Stockholm}
  \country{Sweden}
}
\email{mflierl@kth.se}
% ---------------
%---------------------------------------------------------
\begin{abstract}
\hanwei{
We present a representation learning framework for financial time series forecasting. One challenge of using deep learning models for finance forecasting is the shortage of available training data when using small datasets. Direct trend classification using deep neural networks trained on small datasets is susceptible to the overfitting problem. In this paper, we propose to first learn compact representations from time series data, then use the learned representations to train a simpler model for predicting time series movements. We consider a class-conditioned latent variable model. We train an encoder network to maximize the mutual information between the latent variables and the trend information conditioned on the encoded observed variables. We show that conditional mutual information maximization can be approximated by a contrastive loss. Then, the problem is transformed into a classification task of determining whether two encoded representations are sampled from the same class or not. This is equivalent to performing pairwise comparisons of the training datapoints, and thus, improves the generalization ability of the encoder network. We use deep autoregressive models as our encoder to capture long-term dependencies of the sequence data. Empirical experiments indicate that our proposed method has the potential to advance state-of-the-art performance.}
\end{abstract}
%---------------------------------------------------------
\maketitle
%---------------------------------------------------------
\section{Introduction}
%---------------------------------------------------------
The subject of financial time series forecasting attracts substantial attention as it provides critical information to investors and policymakers. Although the financial time series is well-known for its high volatility, it is possible to predict the future market trend based on previous prices according to the efficient market hypothesis (EMH). The EMH assumes that all information is incorporated into the price and that the market reaches equilibrium through rational decisions of market participants \cite{7911375}. As a result, one could beat the market by identifying the price irregularities \cite{zuckman19}. Specifically, we are interested in predicting binary time series movements such that investors can make profits by taking correct long or short positions of a financial asset. 

Recent progress in financial forecasting focuses on applying deep learning models on time series movement prediction tasks. Since the time series data has the property that it can generate binary labels itself by examining the differences between each timestep, classical supervised learning models can be applied to predict the binary time series movements. Several research works show that the use of state-of-the-art deep learning-based models outperform classical statistical models on prediction performance \cite{chua:19}\cite{cohen:stock}\cite{Anastasia}\cite{Cottrell:17}. 
 
 One challenge of using deep learning models for finance forecasting is the shortage of available training data. For example, consider a dataset that contains daily price information of a stock over a decade. Each year has approximately $252$ trading days. Assume that the size of the time window is $30$ days and the time window moves $1$ day per time. If $80\%$ of the available data is taken as the training dataset, then the total number of available datapoints for training one stock is less than $2000$, i.e., 
%-----------------------
\begin{equation*}
 (252\times10\times80\%-30)/1 + 1 = 1987.
\end{equation*}
%-----------------------
This limited amount of training data can easily result in overfitting of the deep neural network classifiers. 

In this paper, we use representation learning techniques for the forecasting task. The ideal case is that the model can be trained to discover underlying patterns related to the price changes from past temporal instances. The forecasting model then can identify patterns in the current time window that are similar to those in the past and examine how prices reacted. Specifically, we propose a neural network-powered conditional mutual information (CMI) estimator for learning useful representations for predicting price movements. Compared to the correlation coefficient which is a measure of linear dependencies, the CMI can capture non-linear dependencies between variables in conditional settings. 

We leverage the fact that the time series movement prediction task can be self-supervised and incorporate the label information into the representation learning tasks. We consider a class-conditioned latent variable model and set the objective to maximize the mutual information between the latent variables and the label information conditioned on the encoded observed variables. Our proposed CMI estimator naturally prompts pairwise comparisons of variables for determining whether they are sampled from the same class or not. This effectively creates a much larger training dataset and makes the trained model less susceptible to the overfitting problem.

With the proposed CMI estimator, we use a two-step method for time series forecasting. First, we train an encoder neural network to maximize the CMI-based objective function. Then, we feed the trained encoder with the input data and map them into low-dimensional representations. Then, we use the obtained representations for downstream tasks. Since we are interested in binary movement prediction of time series, we use the extracted representations to train a simple binary logistic regression classifier to predict the time series movement.
%---------------------------------------------------------
\section{Contrastive-based Learning}
%---------------------------------------------------------
\subsection{A Class-Conditioned Generative Model}
%---------------------------------------------------------
\hanwei{We consider a generative view for time series forecasting tasks. We use a class-conditioned latent variable model to describe the generative process. First, a latent variable is drawn from a class-conditioned distribution. For time series forecasting, we can view the upward and downward movement as two classes. Then, an observed variable is generated by sampling from an observed data distribution, conditioned on the sampled latent value. 

The objective of the representation learning is to approximate the class-conditioned latent distribution. We can use a neural network powered encoder to parametrize the posterior distribution of the latent variable given the observations. Since the mutual information takes account of nonlinear dependencies and high-order statistics between random variables, the mutual information between the label and the latent variables is a useful criterion for learning representations for classification \cite{Torkkola:00:icml}. It can be shown that the lower bound of the classification error based on the latents is inversely proportional to the value of the mutual information between label and latent variables. Let $X \in \mathcal{X}$ denote the observed variable, $Z \in \mathcal{Z}$ the latent variable, and $Y$ the trend class variable. According to Fano's inequality, we have
%-----------------------
\begin{equation}
p(y \neq \hat{y}) \geq \frac{H(Y|Z)-1}{\log N_y} = \frac{H(Y)-I(Z; Y)-1}{\log N_y}, 
\label{obj:fano_}
\end{equation}
%-----------------------
where $\hat{y}$ is the predicted label and $N_y$ is the number of classes. As a result, the lower bound of the classification error will be minimized when the mutual information $I(Z, Y)$ is maximized. 

Since the latent variables need to be sampled from the approximated distribution which is parametrized by the encoder network, we use the condition mutual information between the latent variable and its class, given the encoded observed variable as criterion for learning the encoders. Let $g_\theta(X) = C$ denote the encoder that maps the input observed variables to the latent space, where $\theta$ denotes the parameters of the encoder. The objective function for learning parameters of the encoder is given as
%-----------------------
\begin{equation}
\label{obj:mi}
\argmax_{\theta}I(Y;Z|g_\theta(X)).
\end{equation}
%-----------------------

%The above variables can form a Markov chain $Z\leftarrow C \leftarrow X \leftrightarrow Y$ such that the label information is incorporated for representation learning. 
%The conditional independence property of the Markov chain model gives $p(x, z|y) = p(x|y)p(z|y)$. 
%-------------------------------------------------
\begin{figure}[!ht]
    \centering
    \includegraphics[width =.47\textwidth]{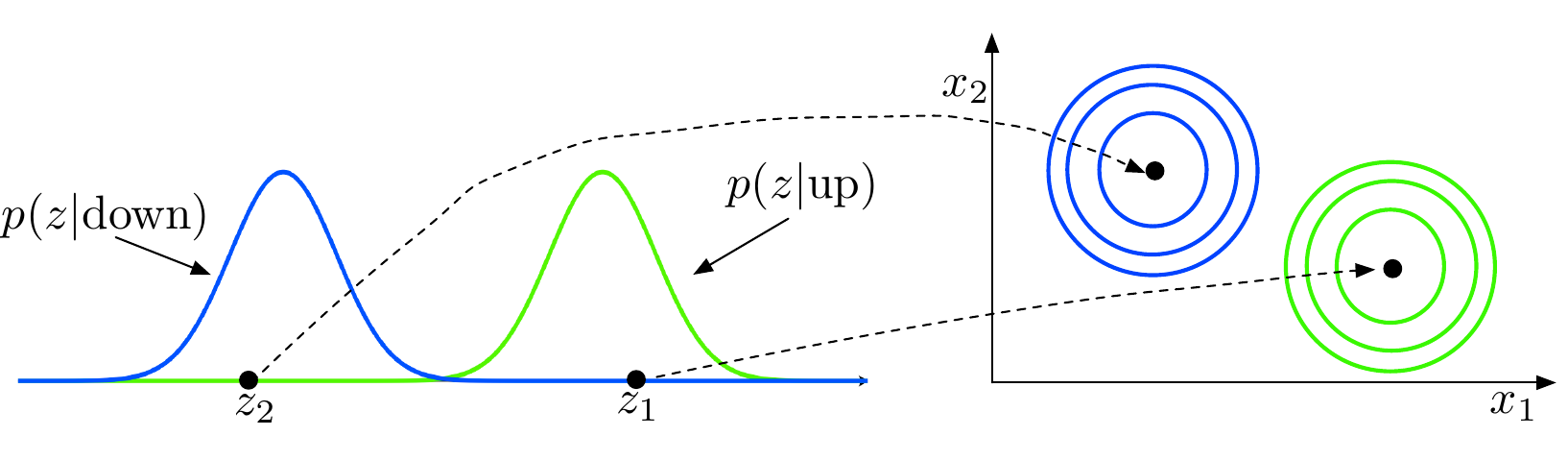}
    \caption{An illustration of the generative process for a two-dimensional data space and a one-dimensional latent space. The blue and green cycles in the right figure indicate the distributions of the observed variables.}
    \label{fig:generative}
\end{figure}}
%-------------------------------------------------
%---------------------------------------------------------
\subsection{CMI Estimator}
%---------------------------------------------------------
%-----------------------
%\begin{equation}
%I(Y; X, Z) \geq I(Y; g_\theta(X), Z) = I(Y;C, Z).
%\end{equation}
%-----------------------
According to the chain rule of mutual information, we can rearrange the expression (\ref{obj:mi}) as
%-----------------------
\begin{align}
I(Y;Z|g_\theta(X)) &= I(Y;Z|C) \\
&= I(Y; C, Z) - I(Y;C) \\
\label{eq:cmi:data_inequality}
&\geq I(Y; C, Z) - I(Y;X) \\
\label{obj:mi_new}
&=I(Y; C, Z)-\log K,
\end{align}
%-----------------------
where $K$ is the number of labels and we assume a uniform prior over the labels such that $I(Y;X) = \log K$. (\ref{eq:cmi:data_inequality}) from the data processing inequality $I(X; Y) \geq I(Y; C)$. Hence, we can use (\ref{obj:mi_new}) as a lower bound of the original objective function of mutual information between the latent variables and the label information conditioned on the encoded variables.

The mutual information between two discrete random is defined as \cite{Cover2006}
%-----------------------
\begin{align}
 I(Y;C,Z) &= \mathbb{E}_{p(y,c,z)}\left[\log \frac{p(c, z, y)}{p(y)p(c, z)}\right] \\
 \label{eq:objexpr}
 &=  \mathbb{E}_{p(c, z, y)} \left[\log\frac{p(c, z|y)}{p(c,z)}\right].
\end{align}
%-----------------------
The conditional mutual information is difficult to compute analytically for high-dimensional vectors. Instead, we can derive a lower bound of the CMI with the help of tractable critic functions to approximate the density ratio $\frac{p(c, z|y)}{p(c, z)}$. Then, the lower bound can be estimated by only having access to the samples of the underlying distributions.

 Ideally, we expect that the encoder can learn a mapping that the datapoints from the same classes are represented together in the latent space. On the other hand, the latent representations belonging to different classes should be dissimilar to each other. Therefore, we choose our critic function to be the exponential function with a similarity measure of $c$ and $z$ as its argument: $f(c, z) = e^{d(c,z)}$, where $d(c, z) = \frac{\langle c, z \rangle}{\|c\|\|z\|}$ and ${\langle \cdot, \cdot \rangle}$ denotes the inner product. We use the normalized inner product to remove the effect of the number of dimensions of the latent vectors and training stability. For our critic function choice, if samples $z$ and $c$ are drawn jointly from the distribution conditioned on a given class, the similarity between $z$ and $c$ should be high and the inner product of $z$ and $c$ tends to be positive, the critic function outputs a score larger than $1$. If samples $z$ and $c$ are from different classes, the $z$ and $c$ should be dissimilar and the inner product of $z$ and $c$ tends to be negative, the critic function outputs a score well below $1$. Then, we have the case that the expectation of $p(c, \bar{z}|y)$ is approximately equal to $p(c, \bar{z})$, where $\bar{z}$ denotes the samples come from a class independent of the given $y$. In this way, the critic function imitates the characteristics of the density ratio $\frac{p(c, z|y)}{p(c, z)}$.
 
Combining the characteristics of the introduced critic function and the expression (\ref{eq:objexpr}), we can derive a lower bound for the original objective function (\ref{obj:mi}) as
%-----------------------
\begingroup
\allowdisplaybreaks
\begin{align}
&I(Y; C, Z)-\log_2(2)\\
=& \mathbb{E}_{p(c, z, y)} \left[\log_2\frac{p(c, z|y)}{2p(c,z)}\right]\\
=& -\mathbb{E}_{p(c,y,z)}\left[\log_2\frac{2p(c,z)}{p(c,z|y)}\right]\\
\geq& -\mathbb{E}_{p(c, y, z)}\left[\log_2\left(f\left(c^{(y)}, z^{(y)}\right)+1\right)\right]\\
\approx &-\mathbb{E}_{p(c, z, y)}\left[\log_2\left(f\left(c^{(y)}, z^{(y)}\right)f\left(c^{(y)}, \bar{z}\right) +1\right)\right]\\
=&\mathbb{E}_{p(c,z,y)} \left[\log_2\frac{1}{f\left(c^{(y)}, z^{(y)}\right)f\left(c^{(y)}, \bar{z}\right)+1}\right]\\
=& \mathbb{E}_{p(c,z,y)} \left[\log_2\frac{f\left(c^{(y)}, z^{(y)}\right)}{f\left(c^{(y)}, z^{(y)}\right) + f\left(c^{(y)}, \bar{z}\right)}\right]\\
\label{equ:critic_rep}
=& \mathbb{E}_{p(c,z,y)}\left[\log_2\frac{e^{d\left(c^{(y)}, z^{(y)}\right)}}{e^{d\left(c^{(y)}, z^{(y)}\right)} + e^{d\left(c^{(y)}, \bar{z}\right)}}\right]
\end{align}
\endgroup
%-----------------------
The introduced critic function not only approximates the density ratio but also relates the estimation of the mutual information to the similarity relation of the latent codes. In our case, the critic function approximates the density ratio by evaluating the cosine similarity between the latent codes. With the introduced critic function, the CMI estimate is maximized when the encoder performs a similarity-preserve mapping.
%---------------------------------------------------------
\subsection{CMI-Based Contrastive Loss}
%---------------------------------------------------------
We can relax the maximization of (\ref{equ:critic_rep}) and cast it as a binary classification problem. We modify the expression inside of the expectation operator of (\ref{equ:critic_rep}) as
%-----------------------
\begin{align}
\hat{ \tilde{y}} = -\log\frac{1}{1 + e^{d\left(c^{(y)}, \bar{z}\right)-d\left(c^{(y)}, z^{(y)}\right)}}.
\label{negative_log_loss} 
\end{align}
%-----------------------
In this way, we can interpret (\ref{negative_log_loss}) as the predicted probability of whether latent pairs of $z$ and $\bar{z}$ are sampled from the same class or not as it in the range of $[0, 1]$ with binary logarithm. Specifically, if $z$ and $\bar{z}$ are samples from the same class, then $\hat{\tilde{y}}$ should be a number close to $1$. If $z$ and $\bar{z}$ are samples from opposite classes, then $\hat{\tilde{y}}$ should be a number close to $0$. %As a result, we can relax the requirement for sampling from opposite classes and let $z$ and $\bar{z}$ being sampled independently instead. 

With this characteristic of (\ref{negative_log_loss}), we can create new labels accordingly. We set new labels $\tilde{y} = 0$ if latent pairs of $z$ and $\bar{z}$ are sampled from the conditional distribution given the same class, and $\tilde{y} = 1$ if $z$ and $\bar{z}$ are sampled from the conditional distributions given opposite classes. Therefore, we can maximize (\ref{equ:critic_rep}) through minimizing a contrastive loss between the predicted labels $\hat{\tilde{y}}$ and the newly generated true labels $\tilde{y}$
%-----------------------
\begin{equation}
\label{equ:loss_function_final}
L_S = \frac{1}{\tilde{N}}\sum_i^{\tilde{N}} \tilde{y}_i  \hat{\tilde{y}}_i+(1-\tilde{y}_i)(1-\hat{\tilde{y}}_i),
\end{equation}
%-----------------------
where $\tilde{N}$ is the number of datapoints. 

We use (\ref{equ:loss_function_final}) as the loss function of training the CMI estimator. Since only the encoder parameters $\theta$ is trainable variables, we are effectively optimizing (\ref{obj:mi}) through (\ref{equ:loss_function_final}). 
%If we have zero loss $L_s = 0$ , then we can recover the maximum mutual information of $1$ bit for binary-class case
% --------------------
%\begin{equation}
%    I(Y;X, Z) = I(Y; C, Z) =  L_S + 1 = 1 \ \text{bit}.
%\end{equation}
% --------------------
% --------------------------------------------------------------------------------------------------
\subsection{Contrastive Sampling Algorithm}
% --------------------------------------------------------------------------------------------------
The CMI estimator can be trained using the the mini-batch gradient descent method. \hanwei{Instead of making prior assumptions about a specific probability family to which the latent distribution belongs, we consider a non-parametric method for maximizing the mutual information. The encoder mapping creates a sample set for the latent variables. The latent variables are directly sampled from the set of encoded variables. For comparison, the classical VAE uses the reparametrization trick to sample from a Gaussian distribution which is parametrized by the encoded variable.} 
 
 For each batch from the original dataset, we create new training samples by iteratively sampling pairs from the original batch and compute new labels through XOR operation $(x_i, x_j,y_i\oplus y_j)$, for $i, j = 1,\cdots,b$. By shuffling the training data after each epoch, the model is exposed to different data pairs during the training. The steps of performing pairwise comparisons for model optimization is summarized in Algorithm \ref{alg:CSA}.
 % --------------------------------------------------------------------------------------------------
\begin{algorithm}[ht]
\begin{algorithmic}
\STATE {\bf Require} Feed the training batches to the encoder and obtain the set of latent codes $S = \left\{c_1^{(y_1)}, \dots, \dots, c_{N}^{(y_N)}\right\}$, where the superscript of an element indicates the sampled class. 
\FORALL{batch} 
\STATE{\bf 1.} For every latent code $c_i$, sample an additional latent code $\left(c_j^{(y_i)},c_k^{(y_k)}\right)$ as the latent variable pair $(z, \bar{z})$ where $i, j, k = 1,\cdots,b$ and $b$ is the batch size. 
\STATE{\bf 2} Create new labels by the XOR operation $\tilde{y} = y_i\oplus y_k$.
\STATE{\bf 3} Compute the predicted probabilities of class labels (\ref{negative_log_loss}).
\STATE{\bf 4} Compute and minimize the contrastive loss (\ref{equ:loss_function_final}) with respect to the encoder parameter $\theta$.
\ENDFOR
\end{algorithmic}
\caption{Contrastive Sampling Algorithm}
 \label{alg:CSA}
\end{algorithm}
% --------------------------------------------------------------------------------------
\subsection{System Model}
\begin{figure}[!ht]
    \centering
    \includegraphics[width =.45\textwidth]{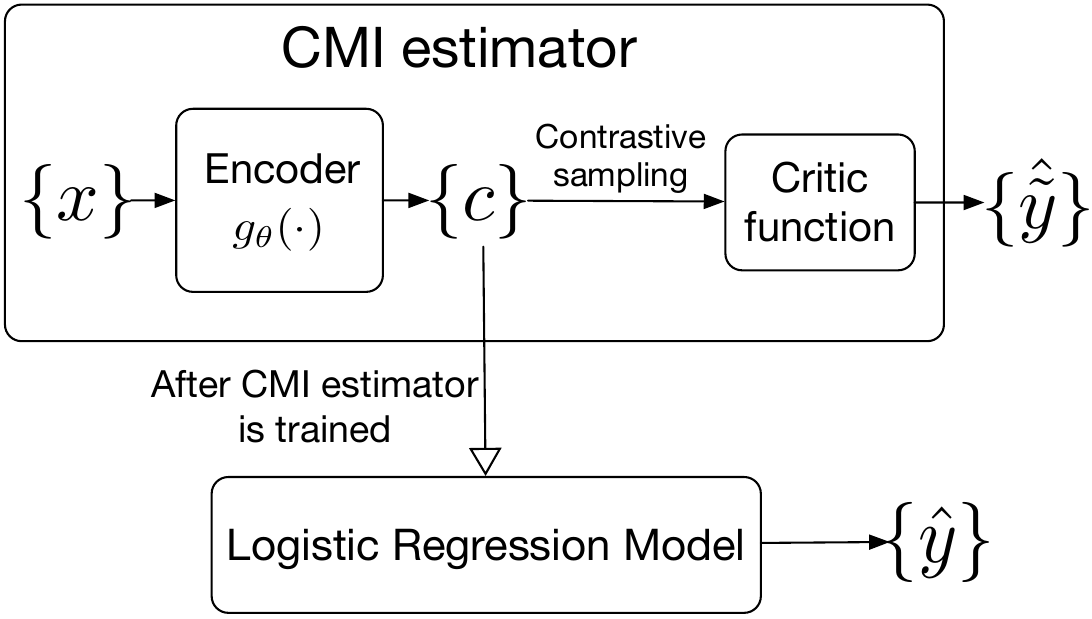}
    \caption{CMI-autoregressive method.}
    \label{fig:cmi}
\end{figure}

\begin{figure}[!ht]
\centering
\begin{subfigure}{.5\textwidth}
\centering
\includegraphics[width =.8\textwidth]{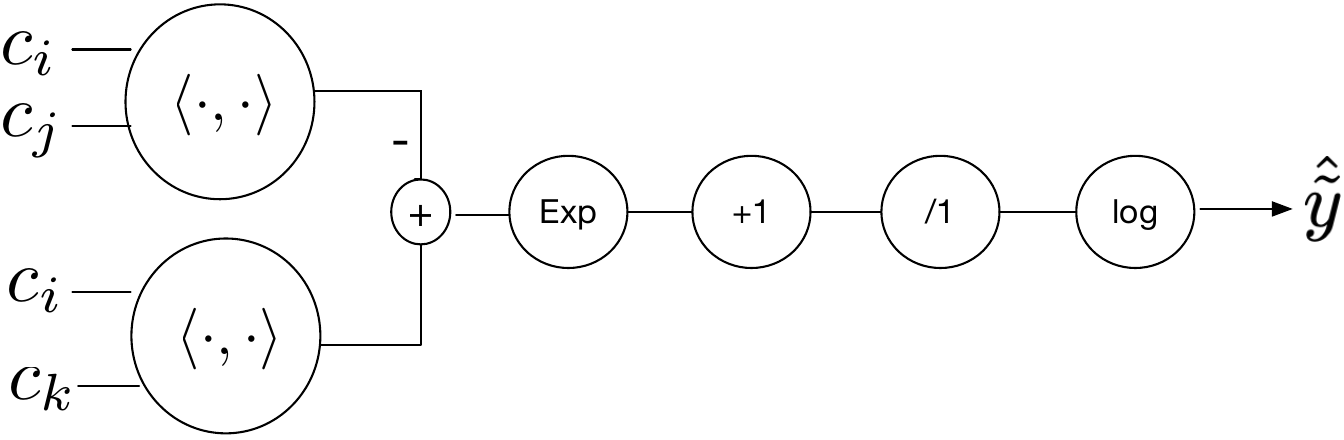}
\end{subfigure}
\begin{subfigure}{.5\textwidth}
\centering
\includegraphics[width =.3\textwidth]{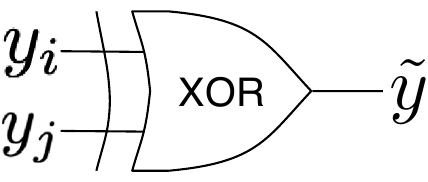}
\end{subfigure}
\caption{Procedures of predicting and creating contrastive-based labels.}
\label{fig:procedure}
\end{figure}

\hanwei{Based on the CMI estimator, we propose a two-step framework for time series movement prediction as shown in Fig. \ref{fig:cmi}. We first use an encoder neural network to map the input to the low-dimensional latent codes. We compute new labels using the obtained latent codes and their corresponding labels according to the contrastive sampling algorithm. A diagram of the contrastive sampling algorithm is shown in Fig. \ref{fig:procedure}. After the encoder is trained based on the contrastive loss, we use the extracted representations to train a simple binary logistic regression classifier to predict the time series movement.}
% --------------------------------------------------------------------------------------------------
% --------------------------------------------------------------------------------------------------
% --------------------------------------------------------------------------------------------------
% --------------------------------------------------------------------------------------------------
% --------------------------------------------------------------------------------------
\section{Comparison With Standard Classification}
% --------------------------------------------------------------------------------------
\subsection{Data Augmentation Perspective}
% -------------------------------------------------------------------------------------
Recent literature such as \cite{recht:17:iclr} and \cite{Arpit:2017:CLM:3305381.3305406} show that the neural networks have the ability of overfitting to the random noise. One approach to prevent the overfitting problem is to use the data augmentation method. However, common data augmentation techniques such as adding noise to the input data may not help with the problem as the stock prices are a low signal-to-noise ratio (SNR) data. 

\cite{Arpit:2017:CLM:3305381.3305406} shows that the convergence time when training the model increases substantially as a function of dataset size. In the limit, the representational capacity of the model is simply insufficient, and memorization is not feasible. As a result, a larger training dataset can prevent the neural networks from memorizing the training dataset. Our proposed CMI estimator naturally prompts pairwise comparisons for determining whether two samples are from the same class or not. This effectively creates much more labeled training data. In this section, we quantitatively compute and compare the amount of information of data-label pairs that contained in the original training dataset and the generated sample pairs. We show that the optimization of the CMI requires the model to encode much more information than learning a binary classifier from the original dataset.

Consider the price movement prediction as a binary classification task. For supervised learning, we simplify the ``memorization'' phenomenon as a process of encoding the information of the data-label assignments into the model. For a given data sample, the amount of information can be encoded into a trained model is $1$ bit
%-----------------------
\begin{equation}
I(x; Y) = H(Y)-H(Y|x) = \log_2(2) = 1 \ \text{bit}.
\end{equation}
%-----------------------
Therefore, the total amount of data-label information can be encoded into a trained model is given by $\sum_{i = 1}^NI(x_i; Y) = N$ bits, where $N$ is the size of the training dataset.

For training the CMI estimators, we create new binary labels for the data pairs $(x_i, x_j)$ datapoints by using exclusive or (XOR) operation $\tilde{y} = y_i\oplus y_j$. Here we set the label $y = 1$ if the movement is upward and $y = -1$ if the movement is downward. As a result, the amount of information can be encoded into a trained model is still $1$ bit
%-----------------------
\begin{equation}
I(x_i,x_j; \tilde{Y}) = H(\tilde{Y})-H(\tilde{Y}|x_i, x_j) = \log_2(2) = 1 \ \text{bit},
\end{equation}
%----------------------- 
 If all of the possible data pairs are used when training the model, the total amount of information can be encoded into a trained model is $\binom{N}{2} = \frac{(N)(N-1)}{2}$ bits. That is, the CMI estimators are trained with approximately up to $\frac{N-1}{2}$ more samples compared to the classical binary classifier. We empirically show that the proposed CMI-based models can reduce the generalization gap between the training and test error in Section \ref{sec:ablation}.
 % --------------------------------------------------------------------------------------------------
 \hanwei{
 \subsection{Representation Learning Perspective}
% --------------------------------------------------------------------------------------------------
Here we compare the CMI estimator with a standard classifier from a representation learning perspective. The standard classifier can be viewed as an encoder that first maps the input datapoint into a representation $c$. Then the classifier computes the similarity scores $d(c, z)$ of $c$ with each \emph{class representation} $z$. This step is typically achieved by adding a linear layer on top of the output of the encoder. The similarity scores are fed into a softmax function to compute the predicted class probabilities. Seen from a representation learning perspective, the encoder of a standard classifier learns the similarity relation between $c$ and $z$ that characterizes the class-conditional probability  
% ------------------------
\begin{equation*}
p(c, z|y) = \frac{e^{d(c, z)}}{Z},
\end{equation*}
% ------------------------
where $Z$ is the normalization constant. 
% ------------------------
\begin{figure}[!ht]
    \centering
    \includegraphics[width =.45\textwidth]{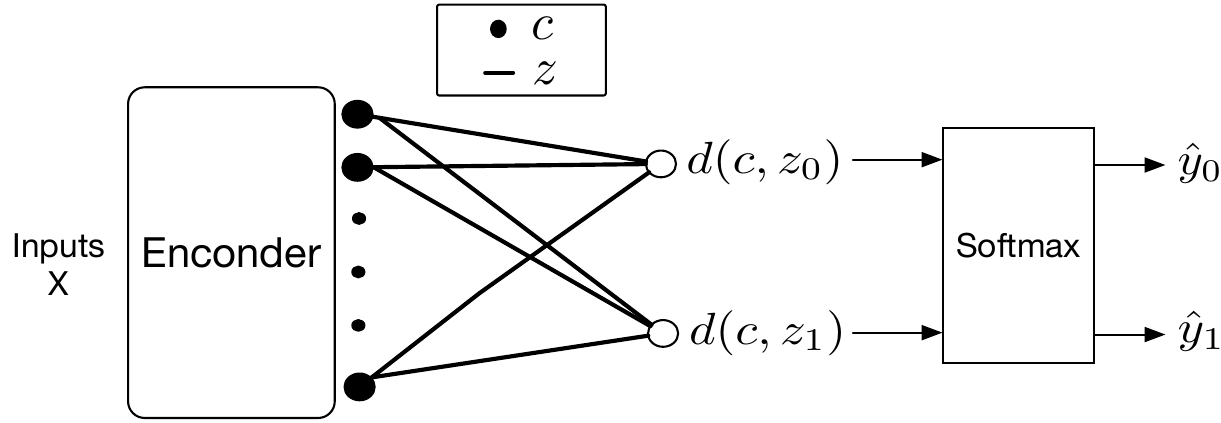}
    \caption{The diagram of a standard classifier. The encoder output representation $c$ is indicated by the black dots. The class representation is the set of weights of the linear layer that connects the encoder output and the softmax function input.}
    \label{fig:direct_classifer}
\end{figure}
% ------------------------

In comparison, our proposed CMI estimator learns the similarity relation between $c$ and $z$ that characterizes the log-density ratio. 
% ------------------------
\begin{equation*}
\log \frac{p(c, z|y)}{p(c, z)} = d(c, z).
\end{equation*}
% ------------------------

The log-density ratio is more reflective of the similarity relation between $c$ and $z$ when compared to the conditional probability distribution. Considering an anomaly detection scenario where only a small fraction of the data is abnormal. The conditional probability learned by a standard classifier can only give the absolute confidence of the predicted classes. On the other hand, the density ratio takes account of the prevalence of both normal and abnormal cases to make the classification decision.}
% ------------------------
\section{CMI Estimator Implementation}
% --------------------------------------------------------------------------------------------------
\subsection{Causal Autoregressive Model}
% --------------------------------------------------------------------------------------------------
We use a deep autoregressive model to encode the input temporal sequences into the latent codes. We use a similar structure as Wavenet\cite{45774}. Specifically, the model uses the so-called dilated casual convolution layers to efficiently capture the long-term dependencies of the time series. The causal convolution preserves the order of the timesteps such that the predictions do not depend on future timesteps. The filter in a dilated convolution layer only applies to every $d$-th entries, where $d$ is the dilation factor. By stocking dilated casual convolution layers, the receptive field of the model grows exponentially and hence captures long-term dependencies. The size of the receptive field is calculated as $r = 2^{l-1}k$, where $l$ is the number of layers and $k$ is the filter size. For example, the model only needs $4$ dilated causal convolution layers to cover $32$ input timesteps. 

Using raw prices as the input data for financial forecasting is not desirable as the prices are subjected to low SNR and increasing for the past two decades generally. Instead, we use the \emph{technical indicators} that derived from the raw price data as the training data. Typically, the technical indicators are summary statistics of timesteps based on its historical data. Hence, the input data for each step is a vector of technical indicators. We regard different input technical indicators as different channels of the input data.
% --------------------------------------------------------------------------------------------------
\subsection{Attention Mechanism}
% --------------------------------------------------------------------------------------------------
We use an attention layer to combine the latent codes of each timestep to create the context vector. We use the context vectors as our learned representations for downstream tasks. Attention mechanism has gained popularity for sequence modeling in recent years \cite{NIPS2017_7181}. The attention mechanism can be viewed as a feed-forward layer that takes the latent codes at each timestep as input and outputs the so-called context vector which is a weighted combination of the input latent codes
\begin{equation}
    \alpha_t = \frac{\exp(a_t^Te_t)}{\sum_{t' = 1}^T\exp(a_{t'}^Te_{t'})}
\end{equation}
\begin{equation}
 c = \sum_{t = 1}^T \alpha_t e_t,
\end{equation}
where $e_t$ is the representation of the $t$-th timestep and $a_t$ is the corresponding weight vector. The attention weight $\alpha_t$ reflects the importance of a timestep of the input to the predicted values of the output. 
% --------------------------------------------------------------------------------------------------
\subsection{Condition on the Stock Identities}
% --------------------------------------------------------------------------------------------------
%-----------------------
In the work of Wavenet, the authors improve the multi-speaker speech generation task by conditioning on speaker identity which is fed into the model in the form of a one-hot vector. Similarly, in our model, we condition on the stock identity information by feeding the one-hot vector of the stock ID to every hidden layer of the encoder
\begin{equation}
h^{l+1} = \text{ReLU}(w^{l} \ast h^{l}+{u^l}^Tv)
\end{equation}
%-----------------------
where $\ast$ denotes the convolution operator, $w^{l}$ is the filter parameters for the $l$-th layer, $h^{l}$ is the output of the $l$-th layer, $v$ is the one-hot vector of the stock identity and $u^l$ is the weights of the $l$-th dense layer. 
% --------------------------------------------------------------------------------------------------
\section{Experiments}
\subsection{Training Data}
% --------------------------------------------------------------------------------------------------
We evaluate our proposed model on stock movement prediction tasks on the \emph{ACL18} dataset \cite{cohen:stock}. ACL18 contains historical price information from Jan-01-2014 to Jan-01-2016 of $88$ high-trade-volume-stocks in NYSE and NASDAQ markets. We use the technical indicators and trend labels that are provided by \cite{chua:19} for the ACL18 dataset. The vectors of technical indicators are obtained by using a sliding window moving along the trading days. That is, one input vector corresponds to one stock per trading day. 

Eleven technical indicators are computed from the ACL18 dataset. Three technical indicators $\Delta_{open_t}$, $\Delta_{high_t}$, $\Delta_{low_t}$ are \emph{interday}-based features. The \emph{interday}-based features can be computed according to the following example $\Delta_{open_t} = open_t/open_{t-1}-1$, where $open_t$ denotes the open price of the stock at timestep $t$. Two technical indicators $\delta_{close_t}$ and $\delta_{adj\_close_t}$ are \emph{intraday}-based features, where $adj\_close$ denotes the adjusted closing price which reflects the stock's value after accounting for any corporate actions, such as stock splits, dividend distribution and rights offerings. The \emph{intraday}-based features can be computed according to the following example $\delta_{close_t} = close_t/close_{t-1}-1$. Other technical indicators are \emph{simple moving average}-based features and are computed as
$\frac{1}{m}\sum_{i = 0}^{m-1}adj\_close_{t-i}/adj\_close_t-1$,
where $m$ is the number of timesteps. We compute six \emph{simple moving average}-based features with $m = 5, 10, 15, 20, 15 ,30$ respectively.

The labels are assigned to each time window based on the next-day movement percent. Movement percent $\geq \% 0.55$ and $\leq -0.5\%$ are labeled as positive and negative examples. We use the time period Jan-01-2014 to Aug-01-2015 as the training dataset and Oct-01-2015 to Jan-01-2016 as the testing dataset.
% --------------------------------------------------------------------------------------------------
\subsection{Model Specification}
% --------------------------------------------------------------------------------------------------
The encoder is constructed by stacking residual blocks. Each residual block consists of a dilated convolution layer with residual and skip connections \cite{7780459}. We set the size of the time window as $64$ trading days. To obtain a receptive field of $64$ timesteps of the encoder, we stack
$6$ residual blocks with the dilation factor doubled for every stacked block. Each dilated convolution layer has $77$ channels with a filter size of $2$. The dimensionality of latents is set to be $96$.

All tested models are trained using the Adam optimizer \cite{kingma:15:iclr} with a learning rate of $1e-4$. The batch size is set to $256$ and the training data is shuffled after each epoch. After the model is trained, the classification accuracy of the learned representation is evaluated by a simple logistic regression model. The decision threshold of the logistic regression is set to be $0.5$. We term our proposed two-step method as CMI-autoregressive method.
% --------------------------------------------------------------------------------------------------
\subsection{Ablation Study}
% --------------------------------------------------------------------------------------------------
\label{sec:ablation}
We evaluate our models on binary classification accuracy and Matthews correlation (MCC). The MCC has a range of $-1$ to $1$ where $-1$ indicates a completely wrong binary classifier,  $0$ indicates a random guess and $1$ indicates a completely correct binary classifier. The MCC takes into account true and false positives and negatives and thus is regarded as a more robust measure of the binary classification quality when the samples of two classes are imbalanced. Due to the low SNR ratio of the stock price data, the accuracy higher than 56$\%$ is generally a satisfying result for binary stock movement prediction \cite{chua:19}.

 We first conduct an ablation study on various components of our model. All the last rows of the result tables are the configurations used in our proposed model and the values are the same. 
% --------------------------------------------------------------------------------------------------
\paragraph{Context Vector Representation}
% --------------------------------------------------------------------------------------------------
We compare different methods of summarizing the latent codes of different timesteps to obtain the final context vector. Specifically, we compare the performance of using max pooling layer, average pooling layer, attention mechanism and the concatenation-dense scheme for obtaining the context vectors. The concatenation-dense scheme first concatenates the latent codes from different timesteps then feed it to a dense layer to obtain the latent vector with the required number of dimensions. In addition, we test the case of directly using the latent codes of the last timestep as the context vector. The result shows that the attention mechanism is critical to the improvement of the performance. 
% --------------------------------------------------------------------------------------------------
\begin{table}[!ht]   %need booktabs package
\scriptsize  
	\caption{Comparison of different context vector representations.}
	\centering
	\begin{tabular}{ccc}
	\toprule
&Accuracy& MCC \\
\hline
	Max pooling  &$51.24\pm 0.5$ &  $0.0112\pm 1e\minus4$ \\
	Average pooling &$52.42\pm 0.5$  &$0.0417 \pm 1e\minus4$  \\
	Concatenation-Dense &$53.92\pm 0.8$ &$0.076 \pm 2e\minus4$ \\
	Last timestep rep. &$51.99\pm 0.4$  &$0.0355 \pm 8e\minus5$ \\
	Attention layer &$56.06\pm 0.7$ &$0.1198 \pm 2e\minus4$ \\
	\bottomrule
	\end{tabular}
\end{table}
% --------------------------------------------------------------------------------------------------
\paragraph{Compare with standard classification} We compare the performance of the deep autoregressive model that is trained to directly classify the prediction movement with our proposed method. The result shows that our proposed method significantly improves the generalization gap (the difference between the training and test error). In practice, a small generalization gap is important as the traders can accurately infer the risk of using the trained model on the unseen data.
% ---------------------------------
\begin{table}[!ht]   %need booktabs package
\scriptsize  
	\caption{Comparison of direct and proposed two-stage classification.}
	\centering
	\begin{tabular}{ccc}
	\toprule
&Test Accuracy& Generalization gap \\
\hline
Vanilla Autoregressive &$50.19$ &  $30.44$\\
CMI-autoregressive&$56.06$  & $2.84$ \\
	\bottomrule
	\end{tabular}
\end{table}
% ---------------------------------
\paragraph{Stock Identity Information} We compare the cases when the model is conditioned on the stock identity information or not. The result shows that the model trained and tested with the stock identity vector has slightly better performance than the model without the stock identity information. 
% ---------------------------------
\begin{table}[!ht]   %need booktabs package
\scriptsize  
	\caption{Comparison of the use of stock identity information.}
	\centering
	\begin{tabular}{ccc}
	\toprule
&Accuracy& MCC \\
\hline
Without stock identity information  &$55.16 \pm 0.5$  &$0.1001 \pm 1e\minus4$\\
With stock identity information &$56.06\pm 0.7$ &$0.1198 \pm 2e\minus4$\\
	\bottomrule
	\end{tabular}
\end{table}
% --------------------------------------------------------------------------------------------------
\subsection{State-of-the-art Comparisons}
% --------------------------------------------------------------------------------------------------
In Table \ref{tab:sfta}, we compare our proposed model with three recently proposed methods for the price movement prediction task. Our proposed CMI-autoregressive method slightly outperforms the reference methods in both classification accuracy and MCC. Note that even a small percentage improvement is appreciated in the financial forecasting task as the investors can exploit the slight statistical advantage by making trades frequently. 
% --------------------------------------------------------------------------------------------------
\begin{table}[!ht]   %need booktabs package
\scriptsize  
	\caption{Next day movement prediction.}
	\centering
	\begin{tabular}{ccc}
	\toprule
&Accuracy& MCC \\
\hline
ALSTM \cite{Cottrell:17}&$54.35\pm 1.0$&$0.0981\pm 3e\minus4$\\
	Stocknet \cite{cohen:stock} &$54.96$&$0.0165$\\
	Adv-LSTM \cite{chua:19} &$55.25 \pm 1.3$ &$0.1055 \pm 5e\minus4$ \\
	CMI-Autoregressive&$56.06\pm 0.7$  &$0.1198 \pm 2e\minus4$\\
	\bottomrule
	\end{tabular}
\label{tab:sfta}
\end{table}
% --------------------------------------------------------------------------------------------------
% -----------------------------------------------------------------------------------------------
\section{Related Works}
% -----------------------------------------------------------------------------------------------
The Long short-term memory (LSTM) model had been a long-established state-of-the-art method for sequence modeling. Hence it becomes a natural choice of majority works for the financial forecasting tasks. Several works improve the vanilla LSTM model by incorporating the attention mechanism. In \cite{Cottrell:17}, the authors propose to incorporate two attention layers to the LSTM encoder-decoder networks. The first attention layer is applied to the input of the model which contains a target price sequence and the related price sequences that are named as \emph{driving sequence}. The attention layer assigns attention weights to each driving sequence for each timestep. The second attention layer is applied to the hidden states across all timesteps output by the encoder. Then, the input of the decoder is a weighted combination of the hidden states of all timesteps. \cite{cohen:stock} proposes to combine the fundamental information extracted from tweets with the technical indicators to predict the binary movements of the stock price. The model first uses an RNN network to extract the corpus embedding from the tweets. Then, the corpus embedding is concatenated with the technical indicators and fed into a variational auto-encoder network. An attention layer is added to the output of the decoder that the prediction of the final timestep is a weighted combination of the predictions of the previous timesteps. \cite{chua:19} introduces an adversarial training method to improve the generalization ability of the attentive LSTM model. The authors propose to introduces small perturbations on the latent codes to simulate the stochasticity of price variable and the perturbed latent samples are termed as \emph{adversarial examples}. The loss function of the model consists of three additive terms: the hinge loss of clean samples, the hinge loss of the perturbed samples and a weights regularization term. As a result, the model is encouraged to correctly classify both clean and adversarial examples. It is shown that the proposed adversarial training method obtains the state-of-art results on the price movement prediction tasks.

Recent advances in audio processing and language modeling have shown the trend of replacing the RNN with autoregressive models on a diverse set of tasks. Although the LSTM has a long-term memory setting, the models trainable by gradient descent cannot have long-term memory in practice. On the other hand, autoregressive models show better training stability and speed compared to the RNN models. In \cite{Anastasia}, the authors use a modified Wavenet model for the financial forecasting tasks. The dilated causal convolutional layers are used to model the temporal target and conditional time series.
  
We note that some of the mentioned works use the mean absolute error (MAE) as their objective function for price predictive tasks. However, the MAE only focuses on minimizing the absolute reconstruction error of the input, it is possible that the reconstructed prices are entirely on the opposite sides of the price movements while achieving a small MAE error. The reconstruction criterion is also problematic as the model can simply learn to copy values from the current prices as the prediction values of the future prices while achieving a small reconstruction error.
  
The estimation of MI is critical to many machine learning applications. However, computing MI between high dimensional continuous random variables is difficult. Recent works use the neural network parameterized variational lower bounds as the MI estimators \cite{pmlr-v80-belghazi18a}\cite{UAI:CCMI}. However, the proposed MI estimators typically suffer from high variance and bias problems when MI is large. \cite{pmlr-v97-poole19a} propose a continuum of lower bounds and can flexibly trades off bias and variance of the MI estimators. \cite{ermon2019_1} propose to clip the density ratio estimates for reducing the variance of the MI estimators. 
Despite the limitations of the current MI estimators mentioned in \cite{tschannen2019}, neural network-powered MI estimators have been applied to various unsupervised representation learning tasks and show state-of-the-art performance \cite{hjelm2019learning}\cite{abs180703748}\cite{abs190509272}\cite{abs190605849}. 
% --------------------------------------------------------------------------------------------------
\section{Conclusions}
Compared to the method of directly training a powerful deep autoregressive model for classifying time series movements, our proposed two-step method can significantly reduce the gap between the training error and the test error. Our proposed CMI estimator transforms the problem to a classification problem whether two samples of the dataset belong to the same class or not. As a result, the CMI estimator is trained with much more data when compared to classic classifiers. The ablation study shows that both the prompted pairwise comparisons and the attention mechanism are critical to the performance improvement. An important direction for future work is to investigate methods for testing the statistical significance of the predictions. 
%A reference implementation is at \url{https://github.com/AlbertOh90/CMI_estimator}.
\bibliographystyle{ACM-Reference-Format}
\bibliography{fine3}
\end{document}